%% file: main.tex
\documentclass[journal]{IEEEtran}
\usepackage{lineno}
\usepackage{subcaption}
\usepackage{algorithm}
\usepackage{algpseudocode}

\usepackage{cite}

\usepackage[pdftex]{graphicx}
\ifCLASSINFOpdf
\else
\fi
\usepackage[cmex10]{amsmath}
\usepackage{hyperref}
\usepackage[capitalize]{cleveref}
\usepackage{amsfonts} 

\usepackage{array}
\usepackage{fixltx2e}

\usepackage{helvet}

\usepackage[normalem]{ulem}

\usepackage{xcolor}

\newcommand{\new}[1]{{\color{blue}#1}}
\newcommand{\newreplace}[2]{{\color{red}\sout{#1}~}{\color{blue}#2}}

\renewcommand{\new}[1]{#1}
\renewcommand{\newreplace}[2]{#2}

\usepackage{booktabs}
\usepackage{multirow}
\usepackage{subcaption}  

\begin{document}
%
\title{SAMSelect: A Spectral Index Search for Marine Debris Visualization using Segment Anything}
%
%
%

\author{Joost van Dalen,
        Yuki M. Asano,
        Marc Rußwurm,~\IEEEmembership{Member,~IEEE}
        
\thanks{Joost van Dalen and Marc Rußwurm are with the Laboratory of Geo-information Science and Remote Sensing, Wageningen University, 6708 Wageningen, the Netherlands. The work was conducted during the Master Thesis of J. van Dalen.}
\thanks{Yuki M. Asano is with the Fundamental AI Lab at the University of Technology Nuremberg, 90461 Nuremberg, Germany.}
}

%
%

\markboth{Accepted by IEEE Geoscience and Remote Sensing Letters}%
{Shell \MakeLowercase{\textit{et al.}}: Bare Demo of IEEEtran.cls for Journals}
%



\maketitle

\begin{abstract}

This work proposes SAMSelect, an algorithm to obtain a salient three-channel visualization for multispectral images. We develop SAMSelect and show its use for marine scientists visually interpreting floating marine debris in Sentinel-2 imagery. These debris are notoriously difficult to visualize due to their compositional heterogeneity in medium-resolution imagery. Out of these difficulties, a visual interpretation of imagery showing marine debris remains a common practice by domain experts, who select bands and spectral indices on a case-by-case basis informed by common practices and heuristics. SAMSelect selects the band or index combination that achieves the best classification accuracy on a small annotated dataset through the Segment Anything Model. Its central assumption is that the three-channel visualization achieves the most accurate segmentation results also provide good visual information for photo-interpretation.
We evaluate SAMSelect in three Sentinel-2 scenes containing generic marine debris in Accra, Ghana, and Durban, South Africa, and deployed plastic targets from the Plastic Litter Project. This reveals the potential of new previously unused band combinations (e.g., a normalized difference index of B8, B2), which demonstrate improved performance compared to literature-based indices. We describe the algorithm in this paper and provide an open-source code repository that will be helpful for domain scientists doing visual photo interpretation, especially in the marine field.
\end{abstract}


%
\IEEEpeerreviewmaketitle

\section{Introduction}
%
%
%
%
\IEEEPARstart{V}{isual inspection} of multispectral remote sensing images is central in many disciplines where integrating domain expertise is essential to provide insight into a particular problem. 
For vegetation monitoring, the interaction between leaf geometry (high NIR reflectance) and photosynthesis (red absorption) can made visible through NIR-red-green false-color band composite (BC) visualizations or vegetation indices like the Normalized Difference Vegetation Index (NDVI), kernel NDVI \cite{camps2021unified} or Enhanced Vegetation Index (EVI) \cite{huete2002overview}. 
In marine applications, the physical properties of floating matter visible in multispectral images are less clear due to the heterogeneity of marine objects. Hence, various spectral indices have been proposed, even in recent years, each targeted towards specific sensors or applications. For instance, algal blooms can be monitored with the Floating Algae Index (FAI)~\cite{hu2009novel}, Normalized Chlorophyll Index~\cite{mishra2012normalized}, or Fluorescence Line Height (FLH)~\cite{hu2017modified}. Non-photosynthetically active objects like marine litter or generic debris are more difficult to identify spectrally due to their compositional heterogeneity in large and mixed medium-resolution pixels. Until recently, specific indices like the Plastic Index~\cite{themistocleous2020investigating}, or Floating Debris Index~\cite{biermann2020finding} have been proposed. Popular land-indices like NDVI are also commonly recommended for marine debris~\cite{biermann2020finding}, even though they lack bio-physical interpretability in the marine context. In these cases, objects can only be distinguishable from water by having an NDVI near zero (no red-edge), while clear marine water typically absorbs more near-infrared than red, resulting in a negative NDVI. Here, NDVI is primarily used as a well-known spectral index that provides reasonable salient visualization in some cases even though the original interpretation of vegetation red-edge is lost.

We postulate that visual interpretation paired with manual trial and error of band visualizations has been a large contributing factor in these recently proposed indices. For instance, the popular Floating Debris Index (FDI) \cite{biermann2020finding} is a modification of the Floating Algae Index (FAI) \cite{hu2009novel} but lacks FAI's physical interpretation of comparing an interpolated near-infrared band with the measured one due to a different scaling factor. Its effectiveness in visualizing floating debris has been demonstrated quantitatively despite its lack of physical interpretation that was present in the original FAI. We believe this practically illustrates our observed trend toward an empirical selection of bands and indices that provide salient visualizations based on manual photo interpretation by domain experts.

Here, this visual inspection by domain experts is approximated by the deep Segment Anything Model (SAM) \cite{kirillov2023segment}, which segments objects-of-interest. SAMSelect uses these segmentations to select the best visualization for the annotations provided by a domain expert.


\section{Method}\label{sec:methods}
\begin{figure*}
  \centering
  \includegraphics[width=\textwidth]{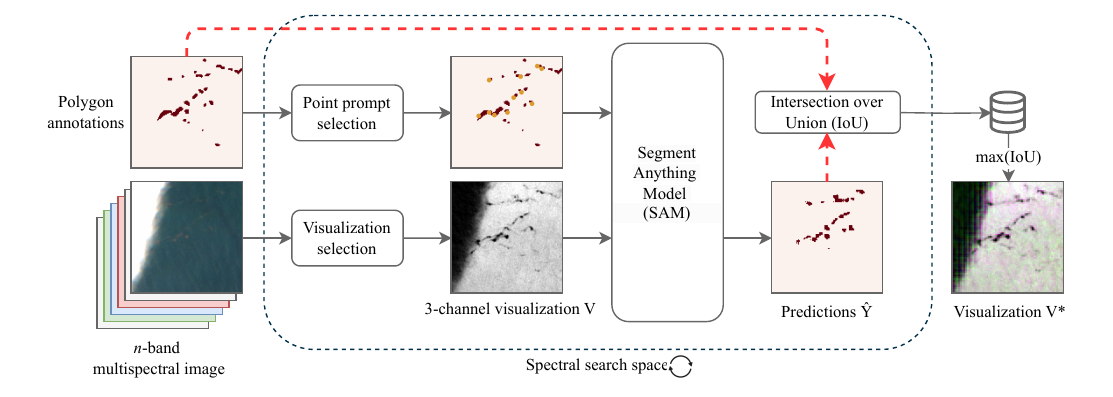}
  \caption{Schematic of the SAMSelect algorithm, automating spectral band selection by maximizing the Intersection over Union (IoU) between SAM-predicted masks and annotated objects. }
  \label{fig:samselect}
\end{figure*}

\noindent
\textit{Process}.
To identify optimal visualizations for marine debris detection, we developed SAMSelect, an automated algorithm that leverages the Segment Anything Model (SAM) \cite{kirillov2023segment} to select the most effective spectral bands or indices. \newreplace{As illustrated in Figure \ref{fig:samselect}, SAMSelect acts as a proxy for human visual interpretation, effectively exploring a vast space for potential visualizations.}{SAMSelect takes in \textit{n}-band multispectral imagery and polygon annotations for objects-of-interest to create a spectral search space based on potential visualizations. It then iterates over the visualizations and point prompts, creating SAM predictions that are evaluated against the annotations. This process, illustrated in Figure \ref{fig:samselect}, allows SAM to act as a proxy for human visual interpretation, effectively exploring a vast space of potential visualizations. }

\newreplace{By providing annotations of specific features of interest, SAMSelect can be tailored to various remote sensing applications. The algorithm can either exhaustively explore all available spectral bands, or focus on a subset defined by user preferences. This flexibility, combined with its data-driven approach, enables SAMSelect to uncover new insights and validate existing spectral indices.}{}

\noindent
\textit{Optimization objective}.
SAMSelect finds a visualization $V : \mathbb{R}^D \to \mathbb{R}^3$ that maps $D$-channel image $\mathbf{X}$ to a 3-channel visual representation where salient features provided in $\mathbf{Y}$ are visible. It searches for a visualization 
\begin{align}\label{eq:samselect}
    V^{\star} = \sum_{\mathbf{X},\mathbf{Y} \in \mathcal{D}} \arg_{V}\max 
        \left[ 
            \text{IoU}
            \left(
                \underbrace{
                    \text{SAM}
                        \left(
                            V(\mathbf{X}), \mathbf{p}
                        \right)
                }_{\hat{\mathbf{Y}}}
            , \mathbf{Y}\right)
        \right]
\end{align}
that maximizes the Intersection over Union (IoU) over SAM prediction of images in a given visualization $\text{SAM}(V(\mathbf{X}), \mathbf{p})$ and their corresponding dense pixel-wise object annotations $\mathbf{Y}$ over a small dataset $\mathcal{D}$ provided by the user. SAM predictions require point prompts $\mathbf{p}$ that we derive from the annotations $\mathbf{Y}$, as detailed later. We implement the maximization by testing an exhaustive search space of possible visualizations.

\noindent
The visualization function V entails two components: 1) a static \emph{histogram normalization} through 1\%-99\% percentile scaling that we determined experimentally in comparison to histogram equalization and min-max scaling in terms of preserving image details and preventing outlier distortion.
2) \emph{band selection} to determine three visualization channels $c_\text{red}$, $c_\text{green}$, $c_\text{blue}$ that we optimize in \cref{eq:samselect}. 
We consider two families of visualizations:

\begin{enumerate}
\item \textit{Band Composites (BC)} assign individual spectral bands to each RGB channel. For example, we can assign different Sentinel-2 spectral bands to these three channels by, for instance, constructing true-color like $c_\text{red} \leftarrow b_\text{B4}$, $c_\text{green} \leftarrow b_\text{B3}$, $c_\text{blue} \leftarrow b_\text{B2}$ or NIR false-color $c_\text{red} \leftarrow b_\text{B8}$, $c_\text{green} \leftarrow b_\text{B4}$, $c_\text{blue} \leftarrow b_\text{B3}$. 
\item \textit{Spectral Index Composites (SIC)} assign different spectral images to the three available visualization bands $c_\text{red} \leftarrow \text{SI}_1$, $c_\text{green} \leftarrow \text{SI}_2$, $c_\text{green} \leftarrow \text{SI}_3$. We consider two general forms of Spectral Indices (SIs): 

\begin{enumerate}
 \item \textit{Normalized Difference Indices (NDIs)} are broadly used for vegetation (NDVI) or surface water (NDWI) and can be expressed generally as
    \begin{equation}\label{eq:ndi}
        \text{$NDI_{b_1,b_2}$} = \frac{(b_1 - b_2)}{(b_1 + b_2)}
    \end{equation}
with two bands $b_1$ and $b_2$. For Sentinel-2, the popular Normalized Difference Vegetation Index (NDVI) can be expressed as $NDVI = \text{$NDI_{B4,B8}$}$ and the Normalized Difference Water Index (NDWI) can be expressed as $\text{$NDI_{B8,B11}$}$ \cite{gao1996ndwi} or \text{$NDI_{B3,B8}$} \cite{mcfeeters1996use}. 

\item \textit{Spectral Shape Indices (SSI)} are used predominantly in marine applications, for instance, as the Floating Algae Index \cite{hu2009novel}, or the Floating Debris Index \cite{biermann2020finding}. In a general form, these indices compare the reflectance at a center band $b_c$ with an linearly interpolated "virtual" band $b^\prime_{c}$ by subtraction $\text{$SSI_{b_\ominus,b_c, b_\oplus}$} = {b_{c} - b^{'}_{c}}$. The linear interpolation involves bands left $b_\ominus$ and right $b_\oplus$ of the center band \begin{equation}
    b^{'}_{c} = {b_\ominus} + ({b_\oplus - b_\ominus})
    \cdot \frac{\lambda_{b_c} - \lambda_{b_\ominus}}{\lambda_{b_\oplus} - \lambda_{b_\ominus}}
\end{equation}
at their wavelengths $\lambda_{b_\ominus}$, $\lambda_{b_c}$, $\lambda_{b_\oplus}$.
For instance, the Floating Algae Index (FAI) using Sentinel-2A data can be expressed as $SSI_{B4, B8, B11} = {B8} - (B4 + ({B11} - B4) \cdot \frac{832.8 \text{nm} - 664.6 \text{nm}}{1613.7 \text{nm} - 664.6 \text{nm}}).$
\end{enumerate}



\end{enumerate}

\noindent
\textit{Point Prompt Selector}.
To guide SAMSelect, prompts can be provided to SAM to specify features of interest within an image. While prompts can vary in complexity (points, bounding boxes, text, or masks) \cite{kirillov2023segment}, we focus on point prompts as they offer a balance between accuracy and ease of use. Point prompts require minimal user effort to generate, making them a practical choice for the application of SAMSelect. We experimented with various prompt configurations using both manually annotated point prompts from the FloatingObjects dataset \cite{ruswurm_large-scale_2023}, and semi-automated approaches relying solely on the provided mask annotations. 

To create semi-automated point prompts, we employed three approaches:
\begin{enumerate}
    \item \textit{Centroids}: \newreplace{For larger masks ($>$10px), we generated centroids based on their geographic distribution.}{For larger litter objects ($>$10px), we generated new point prompts based on the centroid of each feature.} However, this method could sometimes place prompts near the mask edges rather than the true center.
    \item \textit{Skeletonization}: To address this issue, we skeletonized larger masks, reducing them to a single-pixel-wide line that captures the essential structure of the object \cite{zhang_fast_1984}. By randomly sampling a point from this skeleton, we ensured that prompts were placed within the core of the object, rather than at the periphery.
    \item \textit{K-means}: To select representative prompts based on the spectral information of all objects within a patch, we treated prompt selection as a core set problem. \newreplace{A K-means algorithm (K=10) was applied to cluster all pixels based on their spectral features.}{We used the Sentinel-2 bands selected for visualization and applied a K-means algorithm (K=10) to cluster all pixels from the objects based on their spectral features within the selected bands. The K-value was determined empirically: lower K-values resulted in missing more abundant objects within the patch, while higher values caused the model to focus too much on prompted objects, leading to the neglect of unprompted ones.} The pixels closest to the cluster centroids were then selected as prompts, ensuring that they represented the diversity of spectral information of the features.
\end{enumerate}


\textit{Search Space}. Using Sentinel-2 L2A data and conducting an exhaustive search space across all twelve spectral bands, SAMSelect evaluates 1,646 visualizations through the four visualization modules: BC (220), NDI (66), SSI (220), and SIC (1,140). The higher count for SIC reflects its broader range of possible combinations, as it uses the top-10 most informative NDI and SSIs rather than the spectral bands.

\section{Data and Study Sites}\label{sec:data}
We developed and evaluated SAMSelect visualizations on Sentinel-2 imagery with 12 spectral bands \newreplace{on}{ that were all bilinearly interpolated to 10m during the Google Earth Engine data export. We evaluated} three study sites:

\begin{enumerate}
\item Accra, Ghana faces significant waste management challenges, leading to large amounts of plastic debris accumulating on local beaches \cite{dyck_empirical_2016}. Field measurements in 2021 estimated that between 140 to 380 kilograms of plastic are transported daily from the Odaw River into the marine environment \cite{pinto_exploring_2023}. Here, we use Sentinel-2A imagery from the 31st of October, 2018, reveals two distinct types of marine debris: outwash from a waste dump exacerbated by coastal erosion, and large patches of \textit{Sargassum spp}. This site presents a different set of challenges, including distinguishing between plastic debris and naturally occurring \textit{Sargassum}, thus offering a complementary testing to ground to Durban for testing SAMSelect.

\item In Durban, South Africa a severe flooding occurred between the 18th to 22nd of April, 2019. Post-flood social media reports highlighted extensive plastic litter accumulation in the harbor and surrounding marine environment \cite{biermann2020finding}. We use the Sentinel-2B (S2B) imagery from the 24th of April that captures this marine debris, likely washed into the area from the harbor due to the floods. The present cloud and haze coverage makes this site challenging, but highly representative in post-flood scenarios where plastic debris outwash is expected to be high \cite{van_emmerik_impact_2024, cozar_proof_2024}. 

\item The 
Plastic Litter Project (PLP) regularly deploys artificial plastic targets in the Gulf of Gera in Lesbos, Greece \cite{papageorgiou_sentinel-2_2022} for several weeks in a controlled environment. We qualitatively use data from the 2021 campaign where two semi-permanent targets were placed for an extended period to obtain 'pure' Sentinel-2 spectral signatures. We focused on the initial Sentinel-2 overpass with both targets visible, which occurred on June 21, 2021. We use this data, only to test the generalizability of identified visualizations on a verified plastic target. 
\end{enumerate}

SAMSelect requires few annotations to delineate objects-of-interest, which can be provided by hand. In our experiments, we used annotations from the FloatingObjects dataset \cite{ruswurm_large-scale_2023} that contains Sentinel-2 images of Accra, Durban with associated polygon annotations. This dataset includes Sentinel-2 scenes annotated with masks and point locations for visible marine debris. We extracted 128x128 pixel patches centered on these point annotations, corresponding to approximately 1,280 meters on the surface. These patches were carefully selected to maximize debris visibility and minimize overlap between neighboring patches. A total of five patches were extracted for Accra, Ghana, and sixteen for Durban, South Africa. This approach ensured the capture of the most relevant spectral and spatial information of the debris objects while minimizing redundancy from overlapping images.

We applied SAMSelect to Sentinel-2 Level-2A products, excluding only B10 (cirrus) from the spectral search space. To assess the algorithm's performance in finding visualizations, we selected two study sites characterized by unique environmental conditions and substantial marine debris concentrations. These sites, in conjunction with annotations from the FloatingObjects dataset \cite{ruswurm_large-scale_2023}, provide a robust testing environment for SAMSelect.

\section{Results}\label{sec:results}

\begin{figure}
  \includegraphics[width=\linewidth]{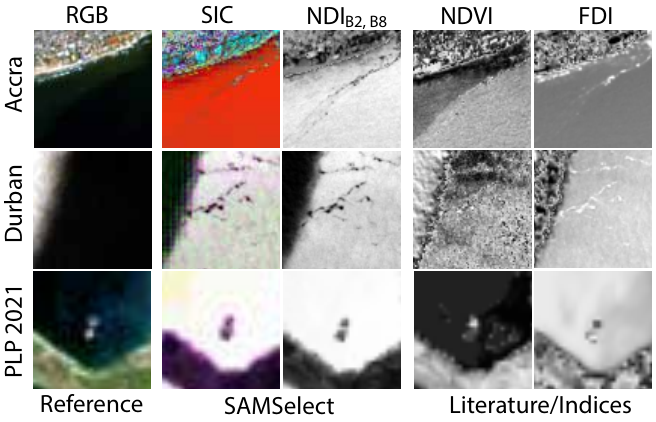}
    \caption{Visualizations of Sentinel-2 images from Accra (2018-10-31), Durban (2019-04-24), and the Plastic Litter Project (2021-06-21). The images include true-color (RGB) and Spectral Index Composites (SIC), as well as the single-channel indices of NDI$_{B2, B8}$,  NDVI, and FDI.}
  \label{fig:combined-patches}
\end{figure}

\noindent
Applying the SAMSelect algorithm yields two different Spectral Index Composites (SIC) for Accra and Durban: $\text{SIC}_\text{Accra}$ is composed of $c_\text{red} \leftarrow \text{NDI}_{B2,B8}$, $c_\text{green} \leftarrow \text{SSI}_{B1,B8,B11}$, $c_\text{blue} \leftarrow \text{SSI}_{B2,B8,B11}$. $\text{SIC}_\text{Durban}$ is composed of $c_\text{red} \leftarrow \text{NDI}_{B2,B8}$, $c_\text{green} \leftarrow \text{NDI}_{B1,B8A}$, $c_\text{blue} \leftarrow \text{NDI}_{B3,B8}$. That SAMSelect yields different spectral index composites is likely due to the varying composition of visible debris both locations. As an easy-to-compute single single indices, we find that the Normalized Difference Index $\text{NDI}_{B2,B8} = \frac{B2-B8}{B2+B8}$ consistently produced most salient visualizations across both Accra, Durban and the PLP2021 targets, as can be see in \cref{fig:combined-patches}. This result is notable given that the blue band is not a commonly used in marine debris detection, but has precedence in a similar  Rotation-Absorption Index (RAI) \cite{loos2012characterization} for oil slick detection. For visualizations on the Plastic Litter Project data, we use $\text{SIC}_\text{Durban}$, as the few pixels of deployed targets are too few to run the SAMSelect algorithm. 


\subsection{Qualitative comparison of visualizations}
In \cref{fig:combined-patches}, we show SAMSelect-derived indices (SIC, NDI$_\text{B2,B8}$) with NDVI and the Floating Debris Index \cite{biermann2020finding}, which are commonly used to interpret floating marine debris. It can be seen that marine debris are difficult to identify in true-color, while they become significantly more distinguishable using spectral indices. We find that NDVI in particular struggles with reliable target identification due to a combination of normalization and limited ability to penetrate the thin haze present in the Durban scene, while the FDI visualizations highlight marine debris. the NDI$_{B2, B8}$ found by SAMSelect provides more pronounced contrast between the target features and the surrounding marine waters. 
This is particularly notable in the Plastic Litter Project 2021 (PLP) targets, which are clearly identifiable with the SAMSelect derived indices, while only one of both targets are clearly delineated in NDVI, FDI. 

\subsection{Quantitative Comparison to Established Indices}

\begin{table}

  \caption{Summary of the best scoring results from the spectral index search space derived from SAMSelect, showcasing four main visualization methods: band composites (BC), normalized difference indices (NDI), spectral shape indices (SSI), and the spectral index composites (SIC). These visualizations are compared against the literature-based indices of NDVI, FDI \cite{biermann2020finding}, and a Principal Component Analysis (PCA).}
    \label{tab:samselect}
    \small\centering
    \begin{tabular}{lll|ll}
    \toprule
        \multicolumn{1}{l}{\textbf{Viz $V$}} & \multicolumn{2}{c}{\textbf{Accra}} & \multicolumn{2}{c}{\textbf{Durban}}  \\
        \cmidrule(lr){2-5}
        \multicolumn{1}{c}{} & Bands & IoU & Bands & IoU \\
        \hline
        NDVI & B8, B4       & 18.7     & B8, B4 & 9.6 \\ 
        FDI  & B6, B8, B11  & 27.7     & B6, B8, B11   & 23.2 \\ 
        PCA  & PC1, PC2, PC3 & 21.3  & PC1, PC2, PC3             & 11.3 \\ 
        \midrule
        NDI & B2, B8 & 36.3     & B2, B8 & 39.5\\
        SSI & B2, B8, B11 & 41.7   & B8A, B9, B11 & 15.1 \\          
        BC & B3, B5, B8A & 36.7    & B3, B8, B8A & 29.6 \\
        SIC & 
        B1,2,8,11
        & \textbf{45.8}    & B1,2,3,8,8A
        & \textbf{42.0}  \\
    \bottomrule
    \end{tabular}
  \end{table}

In \cref{tab:samselect}, we compare the SAMSelect-identified three-channel Spectral Index Composition (SIC) and Band Composite (BC) with the best-performing identified SSI and NDI single-channel indices and compare them to the established NDVI, FDI indices and a Principal Component Analysis Baseline (PCA).
As metric, we report the Intersection over Union (IoU) of the Segment Anything Model \cite{kirillov2023segment} using the visualizations as inputs on annotated polygons of marine debris of the Sentinel-2 scenes in Accra and Durban. 
Overall, the results demonstrate a quantitative improvement over established indices NDVI and FDI, with segmentation scores increasing within a range of 8.6\% to 23\%. In Durban, NDI$_{B2, B8}$ showed even greater improvements, with a 29.9\% increase over NDVI and a 16.3\% increase over FDI, though SSI$_{B2, B8, B11}$ performed less effectively in this region. Comparatively, PCA-based composites produced results similar to NDVI and FDI but had lower segmentation scores than those identified by SAMSelect.
The visualizations found by SAMSelect demonstrated consistent segmentation performance, with differences among the methods ranging from 2.5\% to 9.9\%, except for the subpar performance of SSI in Durban. Generally, assigning individual spectral bands to the RGB channels in BC resulted in similar or decreased segmentation performance compared to single-channel indices like NDI and SSI. This suggests that single-channel indices may be more effective at isolating marine debris features from surrounding water. However, assigning individual NDI and SSI indices to RGB channels using the SIC approach resulted in a slight performance increase over single-channel visualizations, with improvements of 4.1\% in Accra and 2.5\% in Durban. 
Notably, the Spectral Index Composites (SIC) achieved the highest segmentation performance, with scores of 45.8\% in Accra and 42\% in Durban, which is due to the efficient compression of 4 (Accra) and 5 (Durban) spectral bands in three visualization channels.




\section*{Usability in terms of Computational Runtime}

The spectral index search in \cref{eq:samselect} requires repeated model inference over a large search space. Hence, we report here the runtime of the SAMSelect algorithm to highlight it's practical usability in terms of finding a suitable visualization in a reasonable time frame.
The computation time for SAMSelect was evaluated using the Durban dataset, containing 16 annotated images. We assessed two visualization types, being the Band Composites (BCs) and NDIs, on both GPU (NVIDIA GeForce GTX1080) and a laptop CPU. Each visualization type was run three times with an exhaustive search across twelve Sentinel-2 bands, yielding 220 combinations for SIC and 66 for NDI. The reported statistics in Table \ref{tab:sam-runtime}, reflect the average runtime for the exhaustive search, along with the time per combination to support users interested in applying a narrower spectral search space for greater efficiency.
Overall, running SAMSelect requires between 10 minutes and four hours depending on the complexity of visualization in terms of choosing a multi-band SIC or a single-band NDI index. Having access to GPUs greatly accelerates the visualization search by a factor of 7.

\begin{table}
    \centering
    \caption{Average runtime for SAMSelect using GPU and CPU across different visualization types.}    
    \label{tab:sam-runtime}
    
    \begin{tabular}{lcc|cc}
        \toprule
        \textbf{Indices} & \multicolumn{2}{c}{\textbf{Runtime [min]}} & \multicolumn{2}{c}{\textbf{Runtime [sec/comb.]}}\\
        \cmidrule(lr){2-3} \cmidrule(lr){4-5}
        &  GPU & CPU & GPU & CPU \\
        \midrule
        BC         & 34.6 & 237.6  & 9.4 & 64.8   \\
        NDI         & 11.1 & 81.4   & 10.1 & 74.0 \\

        \bottomrule
 \end{tabular}
\end{table}

\section{Discussion and Conclusion}\label{sec:conclusion}
In this letter, we propose the SAMSelect algorithm as a tool that can automatically discover the optimal band combination for visualizing objects of interest in any multispectral scene. By simply providing an image and corresponding annotations, SAMSelect can identify the band combination that maximizes agreement between SAM predictions and the annotations.
In the case of marine debris, SAMSelect has consistently selected the NDI$_{B2, B8}$ combination which outperformed existing indices like NDVI and FDI. This suggests that this combination is particularly effective for distinguishing marine debris from other objects in the scene.
\new{Notably, also visualizations using 20m bands (B5, B8A) achieved high IoUs when paired with at least one 10m band. This was apparent in \cref{tab:samselect} in the Band Composite (BC) for Accra, that makes use of B3 (10m), B5 (20m), and B8A (20m). In these cases, the 20m bands (B5, B8A) provided more spectral context, while the 10m band provides spatial detail.}
While our experiments focused on marine debris detection in Sentinel-2 imagery, SAMSelect is applicable to a wide range of terrestrial and marine applications with other multispectral sensors, such as PlanetScope and Landsat. \new{The SAMSelect can also be applied to much higher-dimensional images, as from hyperspectral sensors or time series. However, for many bands, a more restricted search-space may be required, as exhaustively searching through all possible visualizations will become prohibitively expensive. An alternative is to explore a large search space iteratively with Bayesian Optimization\cite{bayesopt} that prioritizes visualizations in proximity to visualizations that previously scored a high IoU.} 
We provide the source code to SAMSelect at \href{https://github.com/geoJoost/SAMSelect}{https://github.com/geoJoost/SAMSelect}. 
\bibliographystyle{IEEEtran}
\bibliography{references.bib}

%







\input{appendix}


\end{document}

%% file: appendix.tex
\appendix[]
To establish a robust foundation for SAMSelect, a series of ablation experiments were conducted to refine its final configuration. This appendix presents detailed results on the impact of image encoders and atmospheric correction products. The evaluation of point prompt selection, which was a key part of this analysis, is included in the main text under the results section.

\section*{Automated Point Prompt Selection}

To determine the optimal configuration for SAMSelect, we conducted a series of ablation experiments, focusing on point prompt selection as well as image encoders and atmospheric correction products (detailed in the Appendix). The experiments showed that manual annotations consistently outperformed semi-automated methods in terms of segmentation accuracy, with manual annotated prompts achieving the highest IoU score of 29.6\% for the Durban dataset. In contrast, K-means clustering provided a viable alternative when manual annotation is impractical, demonstrating effective performance with a best score of 30.7\% on the Accra dataset. This K-means approach for generating prompts, offers a balance between accuracy and user effort, making it suitable for scenarios where manual annotations are not feasible. While centroid-based and skeleton-based approaches yielded reasonable results in specific cases, they ultimately delivered lower accuracy compared to these two primary approaches.

\begin{table}[htbp]
    \centering
    \caption{Point prompt selection for SAM using manually provided prompts and semi-automated approaches. Different types of prompts are distinguished by SAM to indicate foreground (+), or background objects (-). The evaluation is conducted using the ViT-B encoder on Sentinel-2 L2A imagery, with band composites (BC) of BC$_{B3, B8, B8A}$, and is reported in IoU (\%) scores.}    
    \label{tab:prompts}
    
    \begin{tabular}{lcc|cc}
        \toprule
        \textbf{Prompts} & \multicolumn{2}{c}{\textbf{Accra}} & \multicolumn{2}{c}{\textbf{Durban}}\\
        \cmidrule(lr){2-3} \cmidrule(lr){4-5}
        &  + & + / - & + & + / - \\
        \midrule
        Manual & 20.1 & 17.1    & \textbf{29.6} & 23.2  \\
        K-means         & 26.7  & \textbf{30.7}    & 12.5 & 16.1 \\
        Centroid        & 17.7 & 23.8    & 8.6 & 9.7 \\
        Skeleton        & 17.3 & 21.9   & 12.0 & 12.3 \\
        \bottomrule
 \end{tabular}
\end{table}

\section*{Testing Different Image encoders}
SAM employs three distinct image encoders: ViT-Base (ViT-B), ViT-Large (ViT-L), and ViT-Huge (ViT-H), each offering varying levels of architectural complexity to encode images. As shown in Table \ref{tab:image-encoders}, ViT-B consistently outperformed both ViT-H and ViT-L across our datasets, delivering the highest performance. Interestingly, both ViT-L and ViT-H exhibited lower performance, particularly in the Accra scene, indicating that increasing encoder complexity does not always lead to improved results for all datasets. Moreover, ViT-B is computationally faster than its more complex counterparts, making it a more efficient choice for SAMSelect when dealing with larger spectral search spaces. As a result, ViT-B was selected as the default encoder for SAMSelect.

\begin{table}[htbp]
    \centering
    \caption{Comparative performance of SAM's image encoders using the SAMSelect algorithm. This table displays the best-performing band composites (BC) generated from Sentinel-2 L2A products.}
    \label{tab:image-encoders}
    
    \begin{tabular}{c|ll|ll}
    \toprule
        \multicolumn{1}{c}{\textbf{Encoder}} & \multicolumn{2}{c}{\textbf{Accra}} & \multicolumn{2}{c}{\textbf{Durban}} \\
        \cmidrule(lr){2-5}
        \multicolumn{1}{c}{} & Bands & IoU & Bands & IoU \\
        \hline
       \textbf{ ViT-B} & \textbf{B3, B5, B8A }& \textbf{36.7}    & \textbf{B3, B8, B8A} & \textbf{29.6} \\
        ViT-L & B2, B3, B8 & 5.4      & B3, B8, B8A & 14.4 \\ 
        ViT-H & B3, B5, B6 & 29.1     & B2, B8, B8A & 20.2 \\
    \bottomrule
    \end{tabular}
\end{table}

\section*{Role of Atmospheric correction}
As part of the ablation experiments to configure SAMSelect, we analyzed four atmospherically corrected products: Sen2Cor's \cite{main-knorn_sen2cor_2017} L1C and L2A, and ACOLITE's \cite{vanhellemont_adaptation_2019} L1R and L2R. Results in Table \ref{tab:atmospheric} show that SAMSelect's segmentation performance remained consistent across these products, with differences generally under 1\%. This aligns with \cite{topouzelis_detection_2019}, who found that differences in reflectance values between Sen2Cor and ACOLITE had little impact on floating debris detection. Moreover, SAMSelect's consistent choice of spectral bands further indicates that variations in reflectance do not significantly alter spectral band selection. Nevertheless, for models like SAM, which are not explicitly trained on physical reflectance values, the emphasis is on how well the visualization improves the contrast between marine debris and its surroundings, rather than achieving exact reflectance accuracy.

\begin{table}[htbp]
    \centering
    \caption{Comparative performance of different atmospheric correction products for detecting marine debris using top-of-atmosphere (TOA) and bottom-of-atmosphere (BOA) reflectance values. Products derived from Sen2Cor \cite{main-knorn_sen2cor_2017} are L1C and L2A, while L1R and L2R are processed using ACOLITE \cite{vanhellemont_adaptation_2019}. Results are based on the SAMSelect algorithm using ViT-B with false-color composites (FCC).}
    \label{tab:atmospheric}
    
    \begin{tabular}{l|ll|llc}
    \toprule
        \multicolumn{1}{c}{\textbf{Product}} & \multicolumn{2}{c}{\textbf{Accra}} & \multicolumn{2}{c}{\textbf{Durban}} &  \\
        \cmidrule(lr){2-5}
        \multicolumn{1}{c}{} & Bands & IoU & Bands & IoU & \\
        \hline  
        L1C & B3, B6, B8A & 37.2    & B4, B8, B8A & 29.8 & {\multirow{2}{*}{\rotatebox[origin=c]{270}{TOA}}} \\
        L1R & B3, B6, B8A & 37.2    & B4, B8, B8A & 29.8 &  \\
        \hline 
        L2A & B3, B5, B8A & 36.7    & B3, B8, B8A & 29.6 &  {\multirow{2}{*}{\rotatebox[origin=c]{270}{BOA}}} \\
        L2R & B3, B6, B8A & 36.5    & B4, B8, B8A & 22.5 &  \\
    \bottomrule   
    \end{tabular}
\end{table}